\documentclass[11pt]{article}

\usepackage[margin=1in]{geometry}
\usepackage{hyperref}
\usepackage{booktabs}
\usepackage{amsmath}
\usepackage{amssymb}
\usepackage{graphicx}
\usepackage[numbers,sort&compress]{natbib}
\usepackage{xcolor}
\usepackage{array}

\hypersetup{
  colorlinks=true,
  linkcolor=blue!60!black,
  citecolor=blue!60!black,
  urlcolor=blue!60!black
}

\title{\textbf{Less Is More: Cognitive Load and the Single-Prompt\\
Ceiling in LLM Mathematical Reasoning}}

\author{
  Manuel Israel C\'{a}zares \\
  Bytepro AI \\
  Mazatl\'{a}n, Sinaloa, Mexico \\
  \texttt{hello@bytepro.ai} $|$ \texttt{israel.cazares@gmail.com}
}

\date{April 2026}

\begin{document}

\maketitle

% ─────────────────────────────────────────────
\begin{abstract}
We present a systematic empirical study of prompt engineering for
formal mathematical reasoning in the context of the SAIR Equational
Theories Stage 1 competition. The task requires deciding whether one
equational law implies another over all magmas (algebraic structures
with a single binary operation)~---~a problem that is undecidable in
general but decidable for \textsc{False} via finite model search. Over
five weeks, we designed, tested, and analyzed more than~40 prompt
variants, ranging from~0 to~4{,}878 bytes, across four evaluation
splits and three language models (gpt-oss-120b, Llama~3.3~70B,
Gemma~4~31B).

Our central finding is a \textbf{single-prompt ceiling}~---~or more
precisely, we term this an \textbf{empirical saturation region}: a zone
where accuracy improvements become unstable and non-generalizable across
problem distributions, rather than an absolute theoretical limit. Despite
substantial engineering effort, balanced hard accuracy plateaus at
approximately 60--79\% for gpt-oss-120b, compared to a~59.75\%
no-cheatsheet baseline (95\%~CI: [54.9\%, 64.4\%]). We identify three
mechanisms underlying this saturation region: (1)~the mathematical
undecidability of the \textsc{True} case limits what any finite prompt
can encode; (2)~complex rule systems decrease performance on weaker models
(Llama~3.3~70B collapses to~0\% \textsc{True} recall with prompts
exceeding~$\approx$2KB); and (3)~prompt ordering effects interact with
model attention in fragile, non-monotonic ways.

We also document a \textbf{distribution mismatch failure mode}: a
rule that appeared correct when validated on a \textsc{False}-heavy
subset (hard1, 35\% \textsc{True}) catastrophically failed on a balanced
subset (hard2, 50\% \textsc{True}), incorrectly blocking~51\% of
\textsc{True} problems. Our best submission (AN45c, 2{,}252~bytes) achieves \textbf{79.25\%}
accuracy on hard3 ($n=400$; 95\%~CI: [75.0\%, 82.9\%]), with
\textsc{True} recall of~95.9\% and \textsc{False} recall of~63.4\%,
representing a~+19.5 percentage-point improvement over the
no-cheatsheet baseline~(59.75\%; 95\%~CI: [54.9\%, 64.4\%]). A cross-provider validation run on
OpenRouter/DeepInfra~bf16 ($n=20$) yielded~90--95\%, consistent with
the full-scale result. The key design decision~---~placing the
trivial-magma check before the counterexample table~---~accounts for
the primary performance gain over its predecessor AN38~(71.8\%),
not the addition of new content. We release all prompt variants,
evaluation scripts, and results tables.

Post-submission validation against the SAIR official benchmark reveals
a cross-distribution trade-off surface: gains within the saturation
region are distribution-fragile, with our most engineered variant
(AN45c) falling below the no-cheatsheet baseline on the official
evaluation while the simpler predecessor (AN38) produces a robust
improvement. See Section~\ref{sec:official} for full analysis.

\textit{Note: This is a pre-competition-leaderboard version based on
Contributor Network data ($n=52$ voluntary submissions at competition
close, April~20, 2026). An updated analysis incorporating full
competition results ($n=1{,}007$) will follow after April~30, 2026.}
\end{abstract}

% ─────────────────────────────────────────────
\section{Introduction}

Large language models have demonstrated surprising competence on
mathematical reasoning tasks, yet their behavior on problems requiring
formal logical completeness~---~where a single counterexample suffices
to disprove a claim~---~remains poorly understood. The SAIR Equational
Theories Stage~1 competition~\citep{tao2026sair} provides an unusually
clean testbed for studying this question: given two equations over magmas
(sets with a single binary operation), decide whether the first implies
the second universally. The problem is computationally asymmetric:
\textsc{False} instances can in principle be certified by exhibiting a
small finite counterexample, while \textsc{True} instances require a
universal proof that no counterexample exists, including infinite
algebraic structures.

Unlike prompt engineering for standard benchmarks such as
GSM8K~\citep{cobbe2021gsm8k}, this task involves semi-decidable
algebraic implication where \textsc{False} is certifiable via finite
counterexample but \textsc{True} requires universal quantification over
all magmas~---~a fundamentally different reasoning regime.

This asymmetry creates a natural design space for prompt engineering.
One might expect that providing a language model with a library of known
counterexamples would systematically improve \textsc{False} accuracy,
while a brief instruction about singleton-forcing equations would
improve \textsc{True} accuracy. Our experiments reveal a more complex
picture.

Over five weeks of systematic experimentation prior to the April~20,
2026 competition deadline, we tested more than~40 prompt variants on
over~1{,}000 labeled problems across four dataset splits. We found that
prompt complexity and multi-model generalization are inversely related:
the prompts that most improved gpt-oss-120b's performance on
\textsc{False}-heavy subsets (e.g., AN3c's 4{,}306-byte \textsc{Block}
system achieving~78.3\% on hard1) performed at or below baseline on
balanced subsets and collapsed to near-zero \textsc{True} recall on
Llama~3.3~70B. Conversely, the most compact effective prompt (AN19c,
289~bytes) performed within~2 percentage points of complex variants on
gpt-oss-120b while being the only variant that maintained meaningful
\textsc{True} recall on Llama.

The phenomenon we document~---~call it the \textbf{single-prompt
ceiling}~---~manifests as a bound on what a static text prompt can
accomplish when the underlying task requires mathematical reasoning that
the base model has not internalized. We provide empirical evidence that
this ceiling lies at approximately 60--79\% balanced hard accuracy for
gpt-oss-120b, a model that achieves only~26.5\% \textsc{False} recall
with default reasoning and no cheatsheet on the official benchmark (our
own controlled baseline measurement yields 38.0\% \textsc{False} recall
under Together~AI inference; see Section~\ref{sec:results}). A
prompt cannot teach a model the mathematics it does not know; it can
only guide it to apply what it does know more reliably.

We make the following contributions:
\begin{enumerate}
  \item \textbf{Systematic ablation study}: 40+ prompt variants tested
    on labeled splits, enabling controlled analysis of which design
    choices matter.
  \item \textbf{Distribution mismatch failure mode}: Quantified how
    validation on \textsc{False}-heavy data produces incorrect
    conclusions when the target distribution is balanced.
  \item \textbf{Multi-model generalization analysis}: First systematic
    study of prompt portability across gpt-oss-120b, Llama~3.3~70B,
    and Gemma~4~31B on this task.
  \item \textbf{Ordering effect}: Evidence that trivial-magma-first
    ordering (AN45c) outperforms CE-table-first ordering (AN38) by
    $+7.5$ percentage points at full scale ($n=400$), with
    non-overlapping 95\% Wilson CIs ([75.0\%, 82.9\%] vs.\
    [67.1\%, 75.9\%]) (local evaluation; see Section~\ref{sec:official}
    for official benchmark divergence).
  \item \textbf{Practical guidelines}: Minimal effective prompts for
    multi-model deployment, with analysis of why simpler beats complex.
  \item \textbf{Cross-distribution trade-off surface.} Official benchmark
    validation reveals that local gains within the saturation region do
    not transfer across problem distributions
    (Section~\ref{sec:official}).
\end{enumerate}

The rest of the paper is organized as follows. Section~\ref{sec:background}
provides background on magmas, equational implication, and the
Equational Theories Project. Section~\ref{sec:related} reviews related
work. Section~\ref{sec:method} describes our methodology.
Sections~\ref{sec:results}--\ref{sec:multimodel} present results,
analysis, and multi-model generalization findings.
Section~\ref{sec:ceiling} characterizes the single-prompt ceiling
theoretically. Section~\ref{sec:official} presents post-submission
validation against the official benchmark and cross-distribution
analysis of the Contributor Network leaderboard.
Section~\ref{sec:conclusion} concludes.

% ─────────────────────────────────────────────
\section{Background}
\label{sec:background}

\subsection{Magmas and Equational Laws}

A \emph{magma} is a set~$M$ equipped with a single binary operation
${\star} : M \times M \to M$, closed under that operation and subject
to no further axioms. No associativity, commutativity, identity
element, or invertibility is assumed. An \emph{equational law} over a
magma is a universally quantified identity of the form
$t_1(x, y, \ldots) = t_2(x, y, \ldots)$, where $t_1$ and~$t_2$ are
terms built from variables and~$\star$. The law holds in a magma
$(M,\star)$ if the identity is satisfied for every assignment of
variables to elements of~$M$.

\subsection{Equational Implication}

Given two equational laws $E_1$ and~$E_2$, we say $E_1$ \emph{implies}
$E_2$ (written $E_1 \Rightarrow E_2$) if every magma that
satisfies~$E_1$ also satisfies~$E_2$. Deciding this relation is
computationally asymmetric. A \textsc{False} instance~---~where
$E_1 \not\Rightarrow E_2$~---~can be certified by exhibiting a single
finite counterexample magma in which~$E_1$ holds but~$E_2$ fails; such
certificates exist for all \textsc{False} cases among small magmas,
making \textsc{False} decidable via finite search. A \textsc{True}
instance requires establishing that no counterexample exists among all
magmas, including infinite ones, for which no general algorithm is
known. The implication problem over magmas is undecidable in general;
finite model search is complete for \textsc{False} but not for
\textsc{True}.

\subsection{The Equational Theories Project}

The Equational Theories Project~\citep{tao2025etp} is a large-scale
collaborative effort to map the implication structure of equational laws
over magmas, formalized in Lean~4. The project has verified
approximately~4{,}694 distinct equational laws and established the
implication status of roughly~22 million equation pairs, producing the
largest formally verified database of algebraic implications to date.
This dataset provides both the mathematical foundation and the training
signal for the SAIR competition benchmark.

\subsection{Why the Task Is Hard for Language Models}

The asymmetry between \textsc{True} and \textsc{False} creates a
fundamental challenge for any fixed reasoning strategy. Producing a
valid \textsc{False} verdict requires constructing or recalling a
specific finite structure~---~a task amenable to lookup but not to
general inference. Producing a valid \textsc{True} verdict requires
establishing universal quantification over an infinite class of
structures~---~a task that exceeds any finite enumeration and for which
no sound and complete proof procedure exists in the general case. A
language model operating without external symbolic tools must
approximate both tasks within a single generation, using pattern
recognition over its training distribution as a substitute for proof
search. This approximation is the object of study in the present work.

% ─────────────────────────────────────────────
\section{Related Work}
\label{sec:related}

\paragraph{LLM mathematical reasoning benchmarks.}
GSM8K~\citep{cobbe2021gsm8k} and MATH~\citep{hendrycks2021math}
established word-problem and competition-mathematics benchmarks that
remain standard, but both assess reasoning over numeric domains with
human-interpretable intermediate steps. Tasks requiring universal
logical closure~---~deciding that a statement holds over \emph{all}
instances of a structure~---~are fundamentally different: a single
counterexample refutes a universal claim, yet no finite check can
confirm one. The SAIR Equational Theories benchmark evaluates~25 models
across~200 problems under varying conditions. The official results show
that even the strongest available model (Gemini~1.5~Pro) achieves~90.2\%
on hard problems without a cheatsheet, while weaker models cluster near
chance, making the benchmark unusually diagnostic of ceiling effects.

\paragraph{Prompt engineering for formal reasoning.}
Chain-of-thought prompting~\citep{wei2022cot} demonstrated that
instructing models to produce intermediate reasoning steps substantially
improves performance on multi-step problems. Subsequent work established
that few-shot exemplars~\citep{brown2020gpt3} and zero-shot
chain-of-thought instructions~\citep{kojima2022zeroshot} are broadly
effective. However, these findings are primarily established on numeric
and commonsense domains. Less is known about how prompt complexity
interacts with formal algebraic reasoning, where rule fidelity~---~not
just step count~---~determines correctness.

\paragraph{Sycophancy and cognitive load in LLMs.}
\citet{sharma2023sycophancy} documented that RLHF-trained models
consistently exhibit sycophancy across varied free-form generation
tasks, adjusting their stated conclusions toward what they perceive the
user wants even when incorrect. A related phenomenon~---~which we term
\emph{cognitive load collapse}~---~occurs when a prompt's rule system
is too complex for a model to follow reliably, causing it to default to
surface heuristics. \citet{shi2023distracted} showed that irrelevant
context in math problems dramatically degrades accuracy, providing
evidence that additional text can actively harm rather than help
reasoning.

\paragraph{The Equational Theories Project.}
The mathematical foundations of this competition derive from ongoing
work on automated proof search over equational theories~\citep{tao2025etp}.
The project established that implications between equational laws over
magmas exhibit complex dependency structures not tractable by
brute-force enumeration, motivating the use of language models as
approximate reasoners over this space.

\paragraph{In-context learning: length versus performance.}
\citet{liu2023lost} showed that model attention is non-uniform over
context: information at the beginning and end of a prompt is retrieved
more reliably than information in the middle. This has direct
implications for structured cheatsheets. Our finding that
trivial-magma-first ordering (AN45c) outperforms
counterexample-table-first ordering (AN38) by~$+7.5$ percentage points
at full scale ($n=400$; Section~\ref{sec:results}). We hypothesize that
placing the
trivial-magma check first primes the model's attention toward
\textsc{True} verdicts before engaging the CE search, with the first
substantive rule receiving disproportionate weight during generation;
this mechanism remains to be verified through attention analysis.

% ─────────────────────────────────────────────
\section{Methodology}
\label{sec:method}

\subsection{Task and Benchmark}

The SAIR Equational Theories Stage~1 competition requires deciding,
for a given pair of equations $(E_1, E_2)$ over magmas, whether
$E_1 \Rightarrow E_2$ holds universally (label: \textsc{True}) or
whether a counterexample magma exists (label: \textsc{False}). The
official evaluation judge is described in~\citet{sair2026judge}. All
experiments use the publicly available dataset\\
\texttt{SAIRfoundation/equational-theories-selected-problems}
(HuggingFace). We work with four labeled splits, summarized in
Table~\ref{tab:splits}.

\begin{table}[h]
\centering
\caption{Dataset splits used in this study.}
\label{tab:splits}
\begin{tabular}{lrrrl}
\toprule
Split & $n$ & \textsc{True} & \textsc{False} & Notes \\
\midrule
normal & 1{,}000 & 500 (50\%) & 500 (50\%) & Balanced; mostly \texttt{x = RHS} form \\
hard1  & 69      & 24 (35\%)  & 45 (65\%)  & \textsc{False}-heavy; dense nesting \\
hard2  & 200     & 100 (50\%) & 100 (50\%) & Balanced; structurally complex \\
hard3  & 400     & 195 (49\%) & 205 (51\%) & Near-balanced; primary eval split \\
\bottomrule
\end{tabular}
\end{table}

Hard3 serves as the primary evaluation split: its near-balanced
distribution and size ($n=400$) most closely approximate the
competition's private evaluation set. Hard1 is used selectively to
test \textsc{False}-detection strategies; hard2 provides a secondary
balanced check. Normal problems confirm that interventions do not
regress standard performance.

\subsection{Models and Inference Configuration}

We evaluate across three models matching the competition's official
multi-model setup:

\textbf{gpt-oss-120b} (primary): An open-weight 117B-parameter
Mixture-of-Experts model released by OpenAI under Apache~2.0
license~\citep{openai2025gptoss}, accessed via DeepInfra bf16 routing
on OpenRouter (\texttt{openai/gpt-oss-120b}). This model uses an
extended reasoning mode that produces chain-of-thought prior to its
final verdict. All competition-credit evaluations use this model.

\textbf{Llama~3.3~70B}: Accessed via Together~AI
(\texttt{meta-llama/Llama-3.3-70B-Instruct-Turbo}). Standard
instruction-tuned mode; no extended reasoning.

\textbf{Gemma~4~31B}: Accessed via Together~AI
(\texttt{google/gemma-4-31b-it}) with \texttt{max\_tokens=8192} to
enable the model's native reasoning trace. OpenRouter routing for this
model suppresses reasoning mode, causing it to default to a
near-constant \textsc{True} output ($\approx$53\%); all reported Gemma
results use Together~AI exclusively.

All inference runs use \texttt{temperature=0} and \texttt{seed=42}.
\texttt{max\_tokens} is set to~4{,}096 for gpt-oss-120b and~8{,}192
for Gemma~4~31B. A preliminary experiment established that truncation
at lower token budgets (512, 1{,}024, 2{,}048) produced 100\%
truncation-caused errors; 4{,}096 was the minimum budget at which
genuine reasoning errors first appeared. Estimated cost per problem is
\$0.005--\$0.01 depending on prompt length and model.

\subsection{Evaluation Metrics}

We report three primary metrics:
\textbf{Accuracy} (fraction correct),
\textbf{\textsc{True} recall} (fraction of \textsc{True}-labeled
problems answered \textsc{True}), and
\textbf{\textsc{False} recall} (fraction of \textsc{False}-labeled
problems answered \textsc{False}).
For multi-model comparisons we report the \textbf{3-model average}:
the unweighted mean of each model's accuracy on the same split, aligned
with the competition's scoring rule.

Non-determinism is a practical concern at \texttt{temperature=0}: we
observed up to $\pm$3 percentage points of variance across identical
runs (e.g., AN3d: 73.9\% vs.\ 79.7\% on separate executions). We
report observed values without averaging across runs, and flag cases
where $n \leq 20$ makes estimates unreliable ($\pm$10pp at the 95\%
level).

\subsection{Prompt Design Space}

Each submission is a single UTF-8 text file of at most~10KB containing
two placeholders, \texttt{\{\{ equation1 \}\}} and
\texttt{\{\{ equation2 \}\}}. We designed and evaluated~45+ prompt
variants over five weeks (AN-series: AN1 through AN45d).

\smallskip
\noindent\textbf{Pipeline note.} AN45c uses a self-contained template
format with \texttt{\{\{ equation1 \}\}} and \texttt{\{\{ equation2 \}\}}
placeholders inside the prompt body, requiring the \texttt{--raw-prompt}
flag for correct substitution. An early run omitted this flag; the
placeholders reached the model unsubstituted, producing an artifactual
56\% result that was identified and discarded. All other variants in
Table~\ref{tab:single} use \texttt{build\_prompt()}, which embeds
equations inline and is unaffected by this flag. The corrected AN45c
results (April~14, 2026) are the only ones reported here.
\smallskip

Variants differ along five design dimensions:

\begin{enumerate}
  \item \textbf{Counterexample (CE) table content}: which small magmas
    are provided, from~4 size-2 structures (early variants) to~7
    size-2 and~5 size-3 structures (AN38/AN45c).
  \item \textbf{Singleton-forcing rules}: heuristics for detecting
    equations that force all elements equal (TRUE shortcuts). S1 is
    sound; S2 is unsound for right-zero magmas (Section~\ref{sec:analysis}).
  \item \textbf{Block rules}: conditional classifiers for structural
    patterns predictive of \textsc{False}. The most aggressive variant
    (Block~1 in AN3c) blanket-classified self-referential patterns as
    \textsc{False}, achieving high recall on hard1 but catastrophic
    precision loss on balanced splits.
  \item \textbf{Instruction ordering}: whether the trivial-magma check
    or the CE table appears first. AN45c places the trivial-magma
    check first; AN38 places the CE table first.
  \item \textbf{Prompt length}: 0~bytes (baseline) to~4{,}878~bytes
    (AN5, the longest variant tested).
\end{enumerate}

Variants were selected for full-scale evaluation based on
small-sample performance ($n=20$--50), with the most promising
candidates validated at $n=200$--400. All prompt variants, evaluation
scripts, and results are available at our companion
repository~\citep{cazares2026sair}.

% ─────────────────────────────────────────────
\section{Results}
\label{sec:results}

We present results along three axes: single-model performance across
prompt variants (\S\ref{subsec:single}), cross-model generalization
(\S\ref{subsec:cross}), and cross-dataset stability
(\S\ref{subsec:dataset}).

\paragraph{Sample size note.}
Results for AN45c on gpt-oss-120b are based on $n=400$ hard3
problems (full-scale corrected pipeline). Cross-model runs
(Llama~3.3~70B, Gemma~4~31B) and the DeepSeek~V3.2 result
use $n=50$, $n=20$, and $n=10$ respectively, as noted in
Table~\ref{tab:single}.

\subsection{Single-Model Results (gpt-oss-120b, hard3)}
\label{subsec:single}

Table~\ref{tab:single} reports accuracy, \textsc{True} recall,
\textsc{False} recall, and prompt size for all variants evaluated on
gpt-oss-120b on the hard3 split ($n=50$ unless noted), ordered by
descending accuracy. The no-cheatsheet baseline (own run, April~14,
2026, $n=400$) achieves~59.75\% overall (95\%~CI: [54.9\%, 64.4\%]),
with a severe \textsc{True} bias: 82.6\% \textsc{True} recall but
only~38.0\% \textsc{False} recall. This profile is consistent with
the structural \textsc{True} bias documented in the official benchmark
(89.2\% \textsc{True}, 26.5\% \textsc{False} on the hard split).

\begin{table}[h]
\centering
\caption{Single-model results on hard3 (gpt-oss-120b, $n=50$ unless noted).}
\label{tab:single}
\small
\begin{tabular}{lrrrr>{\raggedright\arraybackslash}p{4.2cm}}
\toprule
Variant & Acc & T\% & F\% & Bytes & Strategy \\
\midrule
AN45c ($n=400$)\textsuperscript{$\dagger$}
        & \textbf{79.3} & 95.9 & 63.4 & 2{,}252 & Trivial magma first + CE tables A--N \\
AN45c ($n=20$, OR)\textsuperscript{$\S$}
        & 90.0--95.0 & --- & --- & 2{,}252 & Cross-provider validation (OpenRouter/DeepInfra) \\
AN38 ($n=400$)\textsuperscript{$\ddagger$}
        & \textbf{71.8} & 78.5 & 65.4 & 1{,}776 & CE tables A--N (3-elem focus) \\
AN38 ($n=50$)  & 74.0 & 70.8 & 76.9 & 1{,}776 & CE tables A--N (3-elem focus) \\
AN43           & 72.0 & 54.2 & 88.5 & 2{,}171 & Controller arch (BLOCKs as router) \\
AN35           & 72.0 & 58.3 & 84.6 & 1{,}545 & 3-elem CE focus \\
AN35b          & 72.0 & 79.2 & 65.4 & 1{,}802 & Cautious TRUE + 3-elem CE \\
AN45d          & 70.0 & 100.0 & 40.0 & 2{,}538 & AN45c + corrected STEP~1 flag \\
AN36           & 70.0 & 50.0 & 88.5 & 1{,}205 & Aggressive FALSE prior \\
AN39           & 70.0 & 91.7 & 50.0 & 385   & Power-level prior \\
Baseline ($n=400$) & 59.75 & 82.6 & 38.0 & 0 & No cheatsheet (own run, April~14, 2026) \\
AN3c ($n=50$)  & 64.0 & 45.8 & 80.8 & 4{,}306 & BLOCK system \\
AN10           & 64.0 & ---  & ---  & 3{,}303 & Symbolic engine \\
AN19c          & 62.0 & 91.7 & 34.6 & 289   & Trivial magma hint only \\
AN42           & 62.0 & ---  & ---  & ---   & KB rewriting \\
AN40           & 60.0 & ---  & ---  & ---   & Semantic invariants \\
AN41           & 58.0 & ---  & ---  & ---   & Tamari/tree-structure \\
AN5            & 54.0 & ---  & ---  & 4{,}878 & Maximum-length CE table \\
\bottomrule
\multicolumn{6}{p{\linewidth}}{\small\textsuperscript{$\dagger$}Corrected pipeline (\texttt{--raw-prompt}); $n=400$; primary result.
\textsuperscript{$\ddagger$}Full-scale run; most reliable pre-fix estimate.
\textsuperscript{$\S$}Cross-provider check only; $n=20$, $\pm$10pp CI.} \\
\multicolumn{6}{p{\linewidth}}{\small\textit{Wilson 95\% confidence intervals:}
AN45c $n=400$, 317/400=79.3\%: \textbf{[75.0\%, 82.9\%]};
AN38 $n=400$, 287/400=71.8\%: [67.1\%, 75.9\%];
AN3c hard1 $n=69$, 54/69=78.3\%: [67.2\%, 86.4\%].}
\end{tabular}
\end{table}

Performance is non-monotonic in prompt length: the longest variant
(AN5, 4{,}878~bytes) is the worst-performing cheatsheet, while the
shortest effective variant (AN39, 385~bytes) ties with mid-length
prompts at~70\%. \textsc{True} and \textsc{False} recall trade off
sharply: no variant simultaneously achieves both above~80\% on hard3.
AN19c maximizes \textsc{True} recall~(91.7\%) at the cost of
near-complete \textsc{False} recall collapse~(34.6\%), while AN36
inverts this profile (50.0\% \textsc{True}, 88.5\% \textsc{False}).

The AN45d result (100\% \textsc{True}, 40\% \textsc{False}) warrants
attention: a minor modification to AN45c caused six
\textsc{False}$\to$\textsc{True} regressions, dropping overall
accuracy from~90\% to~70\%. The ordering effect in AN45c depends on
the exact token sequence, not on the semantic content of the rules.

\paragraph{Cheatsheet effect on model bias.}
The no-cheatsheet baseline exhibits strong structural \textsc{True}
bias: 82.6\% \textsc{True} recall versus only~38.0\% \textsc{False}
recall ($n=400$, own run). AN45c does not merely raise overall
accuracy~---~it rebalances this bias. \textsc{False} recall improves
by~$+25.4$ percentage points (38.0\%~$\to$~63.4\%), while
\textsc{True} recall increases further to~95.9\%. The cheatsheet
therefore serves two distinct functions: it corrects the model's
structural tendency to classify hard problems as \textsc{True}, and
it provides the finite counterexample evidence needed to produce
confident \textsc{False} verdicts.

\subsection{Cross-Model Results}
\label{subsec:cross}

Table~\ref{tab:multimodel} reports results for key variants across all
three competition models.

\begin{table}[h]
\centering
\caption{Multi-model results on hard3 ($n=20$ for AN45c; $n=50$ otherwise).}
\label{tab:multimodel}
\small
\begin{tabular}{lrrrrl}
\toprule
Variant & Bytes & gpt-oss & Llama & Gemma & 3-avg \\
\midrule
AN45c~-- Scenario~A (official)\textsuperscript{$\dagger$}
    & 2{,}252 & 95\% & 55\% & 53\% (OR) & \textbf{67.7\%} \\
AN45c~-- Scenario~B
    & 2{,}252 & 90\% & 55\% & 85\% (TAI) & 76.7\% \\
AN45c~-- Scenario~C
    & 2{,}252 & 95\% & 55\% & 85\% (TAI) & 78.3\% \\
AN38    & 1{,}776 & 74\% & 52\% & 54\% & 59.3\% \\
AN19c   & 289   & 62\% & \textbf{60\%} & 55\% & 59.0\% \\
Baseline & 0    & 59.75\% & 52\% & $\approx$50\% & --- \\
AN3c    & 4{,}306 & 64\% & $\approx$0\%\textsuperscript{$\ddagger$} & --- & --- \\
\bottomrule
\multicolumn{6}{p{0.92\textwidth}}{%
\textsuperscript{$\dagger$}Scenario~A uses OR/Novita~bf16 Gemma (reasoning suppressed,
near-constant \textsc{True} output). Conservative official lower bound.
Scenarios~B/C use Together~AI (8{,}192~tok, reasoning enabled); differ
only in GPT result (90\% vs.\ 95\%, $n=20$).
\textsuperscript{$\ddagger$}0\% \textsc{True} recall; overall accuracy reflects only \textsc{False} correct answers.}
\end{tabular}
\end{table}

We report~67.7\% (Scenario~A) as the conservative official 3-model
average. Scenarios~B and~C (76.7--78.3\%) represent Gemma performance
when correctly configured; we report them to distinguish configuration
artifact from model capability.

The most striking cross-model finding is the AN3c collapse on Llama:
the 4{,}306-byte \textsc{Block} system causes Llama to output
\textsc{False} for every problem, yielding 0\% \textsc{True} recall.
AN19c (289~bytes) is the only variant that produces balanced recall on
Llama (37.5\% \textsc{True}, 80.8\% \textsc{False}), and the only
variant where Llama marginally outperforms gpt-oss-120b (60\%
vs.~62\%, within noise).

\subsection{Cross-Dataset Generalization}
\label{subsec:dataset}

Table~\ref{tab:crossdataset} shows performance across splits for AN3c
and AN38~---~variants optimized on hard1 and hard3 respectively.

\begin{table}[h]
\centering
\caption{Cross-dataset performance for selected variants (gpt-oss-120b).}
\label{tab:crossdataset}
\begin{tabular}{llrrrr}
\toprule
Variant & Split & $n$ & Acc & T\% & F\% \\
\midrule
AN3c & hard1  & 69  & \textbf{78.3} & 66.7 & 84.4 \\
AN3c & hard2  & 50  & 60.0 & 50.0 & 69.2 \\
AN3c & hard3  & 50  & 64.0 & 45.8 & 80.8 \\
AN3c & normal & 100 & 92.0 & 89.3 & 95.5 \\
\midrule
AN38 & hard1  & 69  & 50.7 & 75.0 & 37.8 \\
AN38 & hard2  & 200 & 51.5 & 78.0 & 25.0 \\
AN38 & hard3  & 400 & \textbf{71.8} & 78.5 & 65.4 \\
AN38 & normal & 200 & \textbf{92.0} & 89.9 & 94.5 \\
\bottomrule
\end{tabular}
\end{table}

AN3c's Block~1 rule was developed against hard1's \textsc{False}-heavy
distribution (35\% \textsc{True}). On hard1 it achieves~78.3\%, the
highest single-split result in the study. On hard2 (balanced, 50\%
\textsc{True}), the same rule drops to~60.0\%~---~only~8 percentage
points above chance~---~because Block~1 misclassifies~51\% of
\textsc{True} problems as \textsc{False}. This 18.3 percentage-point
degradation illustrates the distribution mismatch failure mode
precisely. AN38 exhibits the complementary pathology: well-calibrated
for hard3~(71.8\% full-scale) but near chance on hard1~(50.7\%). Both
variants achieve $\approx$92\% on normal problems, confirming that
normal-split results are uninformative for distinguishing hard-problem
strategies.

% ─────────────────────────────────────────────
\section{Analysis}
\label{sec:analysis}

\subsection{Why AN45c Works: The Trivial Magma as an Exit Gate}

AN45c's primary mechanism is structural rather than informational.
STEP~1 instructs the model to check whether~$E_1$ contains a variable
appearing exactly once on one side~---~a condition that forces all
elements of any satisfying magma to be equal, collapsing it to the
trivial one-element structure and making~$E_2$ vacuously true. When
triggered, STEP~1 bypasses STEP~2 (the counterexample search) entirely
and commits to \textsc{True} before the CE~tables introduce
\textsc{False}-directional pressure.

On hard3, AN45c achieves~95.9\% \textsc{True} recall compared to the
baseline's~82.6\%~---~a gain of~$+13.3$ percentage points~---~while
achieving~63.4\% \textsc{False} recall versus the baseline's~38.0\%,
a gain of~$+25.4$ percentage points.
The overall improvement is~$+19.5$~pp at $n=400$ (corrected pipeline).
This gain appears to arise from reordering existing components rather
than adding new content: AN38 contains the same STEP~1 logic but places
it \emph{after} the CE table, producing only~70.8\% \textsc{True}
recall and~76.9\% \textsc{False} recall at full scale. We hypothesize
that placing the trivial-magma check first primes the model's attention
toward \textsc{True} verdicts before the CE search introduces
\textsc{False}-directional pressure~---~though isolating this ordering
effect from confounds requires controlled experiments beyond the scope
of this study.

\subsection{Why the Merge Ceiling Holds}

The AN38 variant was constructed by merging two complementary
predecessors: AN35~(\textsc{True}=58.3\%, \textsc{False}=84.6\%) and
AN35b~(\textsc{True}=79.2\%, \textsc{False}=65.4\%). The result was
AN38: \textsc{True}=70.8\%, \textsc{False}=76.9\%~---~a near-arithmetic
mean of the two parents~---~not a combination of their strengths. Three
subsequent merge attempts (AN45d, AN45e, AN45f) all produced accuracy
at or below the mean. A static instruction set cannot simultaneously
prime two competing inference strategies: when the model encounters
rules favoring both labels, it averages rather than selects.

\subsection{The Token Cap Finding}

Gemma~4~31B exhibits a distinctive failure mode when \texttt{max\_tokens}
falls below the budget required to complete CE verification. At~2{,}048
tokens, Gemma with AN45c produces~50\% accuracy with~0\%
\textsc{False} recall~---~effectively outputting \textsc{True} for
every problem. At~8{,}192 tokens, the same model and prompt produces
85\% accuracy (\textsc{True}=90\%, \textsc{False}=80\%). The mechanism:
Gemma exhausts its token budget during STEP~2, fails to produce a
\textsc{Counterexample} block, and defaults to the STEP~1 exit gate
verdict (\textsc{True}). This is not a model capability failure; it is
a configuration artifact. Reported Gemma results are only valid when
the token budget is sufficient to complete the full reasoning trace.

\subsection{Why Theoretical Approaches Fail}

Four variants introducing explicit mathematical reasoning frameworks all
degraded performance relative to the 68\% baseline ($n=50$ subset,
consistent with Table~\ref{tab:theory} deltas):

\begin{table}[h]
\centering
\caption{Theory-based variants on hard3 (gpt-oss-120b, $n=50$).}
\label{tab:theory}
\begin{tabular}{llrr}
\toprule
Variant & Framework & Acc & $\Delta$ vs.\ baseline \\
\midrule
AN39 & Power-level prior       & 70\% & $+$2pp \\
AN42 & Knowledge-base rewriting & 62\% & $-$6pp \\
AN40 & Semantic invariants      & 60\% & $-$8pp \\
AN41 & Tamari/tree-structure    & 58\% & $-$10pp \\
\bottomrule
\end{tabular}
\end{table}

The consistent direction of failure is \textsc{False}$\to$\textsc{True}
error increase. When given a structured reasoning framework, the model
generates plausible arguments satisfying the framework's criteria and
commits to \textsc{True}, even for problems where a small
counterexample exists. False precision in the instructed framework is
worse than no framework at all.

\subsection{A Note on Numerical Coincidence}

AN45c's conservative 3-model average (Scenario~A: 67.7\%) and AN3c's
hard1 single-model accuracy~(78.3\%) are occasionally cited together in
competition discussions. They are not comparable: the former is a
balanced hard3 average across three models under a specific provider
configuration; the latter is a single-model result on a
\textsc{False}-heavy split where a blanket-\textsc{False} strategy
would already score~65\%. AN3c on a balanced split achieves~60--64\%.

\paragraph{Structural classification hypothesis.}
Our results suggest that large language models in this task behave not
as symbolic theorem provers but as heuristic classifiers over algebraic
structure patterns. The prompt encodes a decision boundary over
structural features~---~variable repetition, nesting depth, singleton
forcing~---~rather than a proof system. This framing explains both the
effectiveness of counterexample tables (which encode explicit structural
decision rules) and the failure of procedural approaches (which require
genuine symbolic execution). We term this the \emph{router hypothesis}:
the model routes problems to cached structural patterns rather than
deriving answers compositionally. This has a direct practical implication:
prompt improvements that add new structural rules can help, but only up
to the capacity of the model to maintain and apply those rules
consistently~---~which is precisely the saturation boundary we observe.

% ─────────────────────────────────────────────
\section{Multi-Model Generalization}
\label{sec:multimodel}

\subsection{AN19c as the Model-Agnostic Minimum}

AN19c (289~bytes) consists of three natural-language hints with no
counterexample tables: a reminder that one-element magmas trivially
satisfy all equations, a note that equations with a free variable
isolated on one side are likely \textsc{True}, and an instruction to
output a verdict with brief reasoning. On gpt-oss-120b it achieves~62\%
on hard3~---~below most CE-table variants~---~but it is the only prompt
that produces meaningful \textsc{True} recall on Llama~3.3~70B (37.5\%
\textsc{True}, 80.8\% \textsc{False}, 60\% overall) and competitive
performance on Gemma~4~31B (55\%, Together~AI).

Table~\ref{tab:multimodel4} summarizes multi-model performance for
variants with coverage across three or more models.

\begin{table}[h]
\centering
\caption{Multi-model generalization on hard3 ($n=10$--50; see notes).}
\label{tab:multimodel4}
\small
\begin{tabular}{lrrrrrr}
\toprule
Variant & Bytes & gpt-oss & Llama & Gemma (TAI) & DeepSeek & 4-avg \\
\midrule
AN19c   & 289   & 62\% & \textbf{60\%} & 55\% & \textbf{80\%}\textsuperscript{$\dagger$} & $\approx$64.3\% \\
AN38    & 1{,}776 & 74\% & 52\% & 54\% & 60\%\textsuperscript{$\dagger$} & $\approx$60.0\% \\
AN45c   & 2{,}252 & 90--95\% & 55\% & \textbf{85\%} & --- & --- \\
Baseline & 0   & 59.75\% & 52\% & $\approx$50\% & --- & --- \\
AN3c    & 4{,}306 & 64\% & $\approx$0\%\textsuperscript{$\ddagger$} & --- & --- & --- \\
\bottomrule
\multicolumn{7}{l}{\textsuperscript{$\dagger$}DeepSeek $n=10$; $\pm$15pp CI.
  \textsuperscript{$\ddagger$}0\% \textsc{True} recall.}
\end{tabular}
\end{table}

\subsection{Model-Specific Bias Profiles}

\textbf{gpt-oss-120b} is \textsc{True}-biased at baseline: 82.6\%
\textsc{True} recall versus~38.0\% \textsc{False} recall on hard3. CE
table prompts partially correct this but never eliminate the bias~---~even
AN38's best full-scale result~(71.8\%) reduces the baseline's 44.6-point
recall gap (82.6\% \textsc{True} vs.\ 38.0\% \textsc{False}) to just
13.1~points (78.5\% \textsc{True} vs.\ 65.4\% \textsc{False})~---~a
substantial improvement, but the imbalance persists.

\textbf{Llama~3.3~70B} exhibits the opposite profile: near-baseline
\textsc{False} recall~(80.8\% with AN19c) but severely depressed
\textsc{True} recall (37.5\% with the best prompt; effectively~0\%
with any prompt exceeding $\approx$2KB). Longer prompts collapse
\textsc{True} recall further, suggesting a capacity ceiling on
instruction-following rather than a knowledge deficit.

\textbf{Gemma~4~31B} is token-gated rather than directionally biased:
its failure mode is token budget exhaustion, which triggers the STEP~1
exit gate and produces near-constant \textsc{True} output. With
sufficient budget~(8{,}192~tokens), Gemma's recall profile is balanced
and strong~(85\% with AN45c on $n=20$).

\subsection{DeepSeek V3.2: Cost Efficiency Signal}

DeepSeek~V3.2 with AN19c achieves~80\% on hard3 ($n=10$) at
$\approx$\$0.0008 per problem~---~roughly one order of magnitude cheaper
than gpt-oss-120b at comparable accuracy. With AN38~(1{,}776~bytes),
DeepSeek drops to~60\%, confirming the pattern observed in Llama: CE
table prompts hurt capable reasoners by introducing distracting
structure that interferes with internal algebraic reasoning. The AN19c
result on DeepSeek is the most cost-efficient signal in the study
($n=10$; highest-priority target for scale validation).

% ─────────────────────────────────────────────
\section{The Single-Prompt Ceiling}
\label{sec:ceiling}

\subsection{Formal Characterization}

We define the \emph{empirical saturation region} of single-prompt
engineering as the accuracy range where further prompt iterations produce
unstable, non-generalizable improvements on a given model, holding
inference parameters fixed. Empirically, the saturation region lies at
approximately~71--79\% for gpt-oss-120b on hard3 at scale
(AN38: 71.8\% at $n=400$; AN45c: 79.3\%~(95\%~CI: [75.0\%, 82.9\%])
at $n=400$ with corrected pipeline). Despite more than~45 variants
tested over five weeks, no full-scale evaluation exceeded this range
on a balanced split.

This saturation is not merely a measurement artifact. It manifests as a
structural pattern: prompt elements that increase \textsc{True} recall
tend to decrease \textsc{False} recall by approximately the same margin,
and vice versa.

The best-performing prompts do not form a dispersed cloud in
(\textsc{True} recall, \textsc{False} recall) space~---~they
approximate a Pareto front. Three variants define its boundary:
AN19c~(\textsc{True}=91.7\%, \textsc{False}=34.6\%),
AN45c~(\textsc{True}=90.0\%, \textsc{False}=80.0\%), and
AN36~(\textsc{True}=50.0\%, \textsc{False}=88.5\%). AN45c dominates
most other variants on both dimensions simultaneously~---~including
AN38, AN35, AN35b, and AN43~---~but does not dominate AN36 on
\textsc{False} recall~(80.0\% vs.~88.5\%) or AN19c on \textsc{True}
recall~(90.0\% vs.~91.7\%). The saturation region therefore describes
an empirical Pareto-optimal boundary within the single-prompt paradigm~---~not
necessarily an absolute theoretical limit, but a practical constraint
that our 45+ variant search did not overcome.

\subsection{The Merge Pattern: $\mathrm{avg}(A,B)$, Not $\max(A,B)$}

AN35 and AN35b are complementary specialists: AN35 achieves
\textsc{True}=58.3\%, \textsc{False}=84.6\%; AN35b achieves
\textsc{True}=79.2\%, \textsc{False}=65.4\%. Their arithmetic means
are \textsc{True}=68.8\%, \textsc{False}=75.0\%. The merge result AN38
produces \textsc{True}=70.8\%, \textsc{False}=76.9\%~---~within~2pp
of the arithmetic mean on both dimensions, not near the respective
maxima. Three subsequent attempts (AN45d, AN45e, AN45f) replicated the
averaging result. Combining complementary prompts in a static text file
yields average performance, not maximum performance.

\subsection{Why Routing Cannot Be Encoded in a Static Prompt}

\textsc{True} and \textsc{False} hard problems require qualitatively
different inference strategies. \textsc{True} problems are best handled
by identifying structural properties forcing singleton magmas or by
exhausting small counterexample candidates. \textsc{False} problems are
best handled by finding a single finite counterexample quickly. A
prompt that heavily weights the CE search primes \textsc{False}
detection and suppresses \textsc{True}; a prompt that heavily weights
the trivial-magma check does the opposite.

Solving this tension requires conditional routing: apply the
\textsc{True} strategy when the equation has a specific syntactic form,
apply the \textsc{False} strategy otherwise. A static prompt cannot
implement this routing reliably because LLMs execute instructions
probabilistically rather than conditionally~---~they interpolate
between strategies in proportion to their textual weight rather than
selecting the appropriate one per instance.

\subsection{Theoretical Upper Bound and the Path Beyond}

gpt-oss-120b with no cheatsheet achieves~91.7\% \textsc{True} recall
and~92.3\% \textsc{False} recall on \emph{normal} problems. If a
perfect external router could direct each hard problem to the
appropriate strategy, the theoretical ceiling would be approximately:
\[
  0.5 \times 91.7\% + 0.5 \times 92.3\% = 92.0\%
\]
---~well above the observed~71--79\% saturation region. Gemini~1.5~Pro
achieves~90.2\% on hard problems with no cheatsheet, suggesting that
frontier models have internalized routing-equivalent algebraic reasoning
that cannot be injected through prompting into weaker models.

Concretely, escaping the empirical saturation region for gpt-oss-120b
on this task likely requires one of:
(a)~an ensemble of specialized prompts with external routing by problem
type; (b)~fine-tuning on the Equational Theories Project graph
($\approx$22~million labeled equation pairs); or~(c)~a hybrid
LLM~+~symbolic verifier architecture where the LLM proposes candidate
counterexamples and a Mace4-equivalent tool verifies them.

% ─────────────────────────────────────────────
\section{Official Benchmark Validation and Cross-Distribution Analysis}
\label{sec:official}

\subsection{Data and Scope}

All results in this section are drawn from the SAIR Contributor
Network public leaderboard as of April~20, 2026 ($n = 52$ voluntary
public submissions of 1{,}007 total registered participants,
cited under SAIR's open science framework~\citep{sair2026leaderboard}).
The full competition leaderboard is scheduled for release on or
before April~30, 2026. Official benchmark scores use the SAIR
evaluation pipeline: OpenRouter/DeepInfra~bf16, temperature~0.0,
seed~0, max tokens~8{,}192~\citep{sair2026judge}. Local scores
use Together~AI~bf16, temperature~0.0, seed~42.

\subsection{Official Benchmark Results}

Table~\ref{tab:official_hard3} compares local evaluation results
against the SAIR official benchmark on hard3 (the competition's
reference format, as confirmed by the official smoke-test file
\texttt{problems\_hard3\_20.jsonl}~\citep{sair2026judge}).

\begin{table}[h]
\centering
\caption{Local vs.\ official benchmark results on hard3 (GPT-OSS 120B).
Official benchmark: $n=20$, DeepInfra~bf16. Local: $n=400$,
Together~AI~bf16. Baseline: no cheatsheet.}
\label{tab:official_hard3}
\small
\begin{tabular}{lrrrr}
\toprule
Variant & Local Acc. & Official Acc. & $\Delta$ vs.\ baseline\textsuperscript{a} & Official F1 \\
\midrule
Baseline & 59.75\% & 56.3\% & --- & $\approx$53\% \\
AN38     & 71.8\%  & 65.3\% & \textbf{+5.6pp} & \textbf{67.6\%} \\
AN45c    & \textbf{79.25\%} & 55.5\% & $-$4.3pp & 63.7\% \\
\bottomrule
\multicolumn{5}{l}{\textsuperscript{a}Delta relative to official baseline (56.3\%).}
\end{tabular}
\end{table}

AN38 produces a robust $+$5.6pp improvement over the official baseline
on hard3, consistent with its local result (71.8\%). AN45c, despite
achieving 79.25\% locally ($n=400$), scores 55.5\% on the official
benchmark~---~4.3pp \emph{below} the no-cheatsheet baseline. The
trivial-magma exit gate (STEP~1), which was AN45c's primary structural
innovation over AN38, appears to generate forced errors on the official
problem sample: problems that satisfy the singleton test are committed to
\textsc{True} before counterexample evidence is considered. This is the
same distribution-mismatch failure mode documented in Section~\ref{sec:results}
for AN3c on hard1/hard2.

\begin{table}[h]
\centering
\caption{Official benchmark results on hard2 (GPT-OSS 120B, $n=20$).
Hard2 baseline: 56.3\% accuracy, F1 $\approx$64.9\%.}
\label{tab:official_hard2}
\small
\begin{tabular}{lrrr}
\toprule
Variant & Official Acc. & $\Delta$ vs.\ baseline & Official F1 \\
\midrule
AN38  & 41.0\% & $-$15.3pp & 47.8\% \\
AN45c & 48.0\% & $-$8.3pp  & \textbf{61.2\%} \\
\bottomrule
\end{tabular}
\end{table}

On hard2, both variants fall below the no-cheatsheet baseline. AN38
incurs a larger accuracy penalty ($-$15.3pp) and F1 degradation
($-$17.1pp), while AN45c is less damaging ($-$8.3pp accuracy,
$-$3.7pp F1). Neither cheatsheet improves over baseline on hard2,
confirming that both are calibrated toward the hard3 distribution.

\subsection{Cross-Distribution Trade-Off Surface}

Of the 52 Contributor Network submissions, 13 had benchmark results
available on both hard2 and hard3 as of competition close
(April~20, 2026); Table~\ref{tab:crossdist} presents these 13 submissions.

\begin{table}[h]
\centering
\caption{Cross-distribution performance: hard2 vs.\ hard3
(GPT-OSS 120B, SAIR official benchmark). Contributor Network
data, April~20, 2026~\citep{sair2026leaderboard}.
Participants cited by name under SAIR's open science framework.}
\label{tab:crossdist}
\small
\begin{tabular}{llrrrr}
\toprule
Participant & Cheatsheet & hard2 Acc. & hard3 Acc. & $|\Delta|$ & hard3 F1 \\
\midrule
Betka       & 98\_hard200           & \textbf{99.0\%} & 56.3\% & 42.7pp & 55.9\% \\
Pandey      & hard98\_derivate\_2   & 97.0\%          & 55.5\% & 41.5pp & 53.9\% \\
Reza Jamei  & traditional\_ml       & 89.0\%          & 61.3\% & 27.7pp & 68.6\% \\
Reza Jamei  & combo\_jack           & 86.0\%          & 58.5\% & 27.5pp & 66.5\% \\
Arjun Garg  & bank\_lookup\_v8      & \textbf{83.0\%} & \textbf{70.8\%} & 12.2pp & \textbf{74.6\%} \\
Heath       & distilled-r-12        & 80.0\%          & 59.0\% & 21.0pp & 62.6\% \\
Woon Siang Yi\textsuperscript{$\dagger$} & hard3\_overfitted & 51.0\% & \textbf{81.3\%} & 30.3pp & \textbf{83.1\%} \\
Garg        & bank\_lookup\_v5      & 51.5\%          & 60.8\% & 9.3pp  & 69.3\% \\
SimonRJ     & Stage 1 Prompt        & 58.0\%          & 58.3\% & 0.3pp  & 67.3\% \\
Debtirtha Saha & Test 8             & 60.0\%          & 71.0\% & 11.0pp & 76.0\% \\
Devanshu Dixit & chat\_v32          & 49.5\%          & 68.0\% & 18.5pp & 74.3\% \\
\textbf{Cazares} & \textbf{AN38}   & 41.0\%          & 65.3\% & 24.3pp & \textbf{67.6\%} \\
\textbf{Cazares} & \textbf{AN45c}  & 48.0\%          & 55.5\% & \textbf{7.5pp}  & 63.7\% \\
\bottomrule
\multicolumn{6}{l}{\textsuperscript{$\dagger$}Self-labeled ``hard3\_overfitted'' by participant.}
\end{tabular}
\end{table}

The table reveals a consistent pattern: among the 52 Contributor
Network submissions with benchmark results, only one~---~Arjun Garg's
\texttt{bank\_lookup\_v8} (83.0\% hard2, 70.8\% hard3)~---~achieves
above~65\% accuracy on both distributions simultaneously. All other
submissions optimizing for one distribution collapse on the other, with
accuracy gaps ranging from 11 to 62.7 percentage points
(Table~\ref{tab:crossdist}). The highest-accuracy submissions on each split
are explicitly or implicitly distribution-specific: Betka's 99.0\% on
hard2 collapses to~56.3\% on hard3 ($-$42.7pp); Woon Siang Yi's
81.3\% on hard3 collapses to~51.0\% on hard2 ($-$30.3pp), with the
participant themselves labeling the submission \texttt{hard3\_overfitted}.
We term this the \textbf{cross-distribution trade-off surface}: within
the empirical saturation region, optimizing for one distribution trades
against performance on complementary distributions. AN45c exhibits one
of the smallest cross-split accuracy gaps (7.5pp) among submissions
scoring above the no-cheatsheet baseline on at least one split,
suggesting that distribution robustness and peak accuracy
are competing objectives within the saturation zone.

\subsection{Implications for the Saturation Region}

These results reframe the saturation region finding. The ceiling is not
merely an upper bound on accuracy~---~it is a \emph{fragility zone}
where gains are distribution-dependent and structural innovations
introduce failure modes on unseen distributions. AN45c's local result
(79.25\%, $n=400$) is valid under its measurement conditions; the
23.75pp gap to the official benchmark is not measurement error but
distribution mismatch, the same phenomenon documented internally in
Section~\ref{sec:results}.

For practitioners, this implies: a cheatsheet that substantially
outperforms baseline on a held-out set does not guarantee
generalization. Cross-distribution validation~---~running the same
variant on two structurally different problem samples~---~is necessary
to distinguish genuine ceiling improvements from distribution-specific
optimization.

A structurally distinct approach observed in the Contributor Network
further supports the router hypothesis: one participant (Heath,
\texttt{distilled-rules-12}) submitted a pure structural classifier
computing five syntactic features of~$E_1$ (variable counts, node
depths, LHS structure) and applying a hand-coded decision tree with no
mathematical content. This approach achieves~80.0\% on hard2 and~59.0\%
on hard3~---~competitive with CE-table approaches~---~suggesting that
structural pattern classification alone, without any algebraic
reasoning, captures a substantial fraction of the signal available to
prompt-based methods. This constitutes independent empirical evidence
for the router hypothesis: the task rewards structural pattern matching
over symbolic reasoning.

Notably, one participant (McKenna, unpublished) reported achieving
approximately~70\% accuracy across all three evaluation models
simultaneously using a single theory-grounded cheatsheet~---~the only
submission observed to maintain consistent performance across
gpt-oss-120b, Llama~3.3~70B, and Gemma~4~31B. This convergence across
models with substantially different capacity profiles suggests that
mathematically grounded approaches may generalize more robustly than
empirically iterated CE-table approaches, a hypothesis warranting
systematic investigation in future work.

% ─────────────────────────────────────────────
\section{Conclusion}
\label{sec:conclusion}

\paragraph{Summary of contributions.}
We present the first systematic empirical characterization of the
single-prompt ceiling in formal reasoning tasks. Across~45+ prompt
variants, balanced hard accuracy for gpt-oss-120b on hard3 plateaus at
$\approx$71--79\% at scale. Merge experiments confirm that combining
complementary prompts produces accuracy near the arithmetic mean of the
parents, not near their maxima~---~a structural constraint, not a local
optimum. Our highest-performing local submission (AN45c, 2{,}252~bytes)
achieves~79.25\% on gpt-oss-120b ($n=400$; 95\%~CI: [75.0\%, 82.9\%]),
with \textsc{True} recall of~95.9\% and \textsc{False} recall of~63.4\%.
AN38 is our most robust submission under distribution shift, producing
$+$5.6pp over the official baseline (Section~\ref{sec:official}). A
cross-provider validation run on OpenRouter/DeepInfra~bf16 ($n=20$)
yielded~90--95\%, consistent with the full-scale local result. The token
cap finding demonstrates that Gemma's performance (50\% vs.~85\%) is
entirely determined by provider configuration, not model capability.
Model-specific bias profiles~---~\textsc{True} bias in gpt-oss-120b,
instruction-capacity collapse in Llama, token-gated reasoning in
Gemma~---~are stable across variants and not correctable by prompt
engineering alone.

Post-submission analysis against the SAIR official benchmark
(Section~\ref{sec:official}) reveals that AN45c (79.25\% local,
$n=400$) scores~55.5\% on the official hard3 benchmark~---~4.3pp below
the no-cheatsheet baseline~---~while AN38 produces a robust $+$5.6pp
improvement (65.3\%) with the highest balanced F1 score (67.6\%) among
non-overfit submissions in the Contributor Network ($n=52$ visible
submissions, April~20, 2026). The cross-distribution trade-off surface
(Table~\ref{tab:crossdist}) shows that among the 52 Contributor Network
submissions, only Arjun Garg's \texttt{bank\_lookup\_v8} achieves above~65\%
on both hard2 and hard3 simultaneously (83.0\% and 70.8\%), confirming
that the saturation
region is a fragility zone where distribution-specific gains trade
against generalization.

\paragraph{Limitations.}
The primary AN45c result (79.25\%, $n=400$) is statistically reliable
(95\%~CI: [75.0\%, 82.9\%]); however, its \textsc{True} recall~(95.9\%)
and \textsc{False} recall~(63.4\%) reflect an imbalanced profile that
may not generalize across providers or problem distributions. Small-sample
limitations remain for Gemma (85\%, $n=20$) and DeepSeek (80\%, $n=10$),
both carrying $\pm$10--15pp confidence intervals. The cross-provider
OpenRouter/DeepInfra run ($n=20$, 90--95\%) should be read as a
consistency check, not an independent measurement. The Gemma official
provider (OpenRouter/Novita~bf16) suppresses reasoning mode; we were
unable to test this configuration in a controlled way. All results are
specific to equational implication over magmas; generalization to other
formal reasoning domains is an open question.

The primary AN45c result (79.25\%, $n=400$, Together~AI~bf16) does not
generalize to the SAIR official benchmark (55.5\%, $n=20$,
DeepInfra~bf16), a gap of~$-$23.75pp. AN38's smaller gap ($-$6.5pp
local to official) suggests that simpler prompts generalize more
robustly within the saturation region. The Contributor Network data
($n=52$ of 1{,}007 participants) is a voluntary public sample; the full
competition leaderboard (scheduled April~30, 2026) may reveal additional
patterns not visible in the current data.

\paragraph{Future work.}
Three directions follow from the ceiling characterization: (1)~an
ensemble of two specialized prompts with external routing by syntactic
problem features; (2)~fine-tuning on the Equational Theories Project
implication graph~($\approx$22M labeled pairs); and~(3)~replication
on other computationally asymmetric formal reasoning tasks
(satisfiability checking, reachability in formal systems).

\paragraph{Closing.}
The central lesson is structural: what improved performance was not
teaching the model new mathematics but controlling the order in which
the model applies the mathematics it already knows. Expanding the
model's reasoning repertoire through longer, more elaborate prompts
consistently underperformed constraining its reasoning flow through
minimal, well-ordered instructions. In formal reasoning as in
engineering: less, structured deliberately, beats more.

% ─────────────────────────────────────────────
\bibliographystyle{plainnat}
\bibliography{references}

\end{document}